\documentclass[10pt,twocolumn,letterpaper]{article}

\usepackage[pagenumbers]{cvpr}

\usepackage{times}
\usepackage{latexsym}

\usepackage[T1]{fontenc}

\usepackage[utf8]{inputenc}

\usepackage{microtype}

\usepackage{inconsolata}

\usepackage{graphicx}

\usepackage{algorithm}
\usepackage{algorithmic}
\usepackage{amsmath}
\usepackage{adjustbox}
\usepackage{xspace}
\usepackage{amsfonts}
\usepackage{booktabs}
\usepackage{multirow}
\usepackage[dvipsnames]{xcolor}

\definecolor{cvprblue}{rgb}{0.21,0.49,0.74}
\usepackage[pagebackref,breaklinks,colorlinks,citecolor=cvprblue]{hyperref}
\makeatletter 
\DeclareRobustCommand\onedot{\futurelet\@let@token\@onedot} 
\def\@onedot{\ifx\@let@token.\else.\null\fi\xspace}
\def\eg{\emph{e.g}\onedot} 
 
\def\ie{\emph{i.e}\onedot}

\makeatother

\title{CineEvolve: An Agentic Self-Evolution Framework for \\ Cinematic Multi-Shot Video Generation}

\author{
    $\textrm{Jingzhe Zhang}^{\#1} \quad \textrm{Peixuan Zhang}^{\#1,5} \quad \textrm{Genying Li}^{1} \quad \textrm{Bing Ma}^{5} \quad \textrm{Shuchen Weng}^{2,4} \quad \textrm{Si Li}^{*1,6} \quad  \textrm{Boxin Shi}^{2,3}$\\
    \small
    $\textrm{}^{1}$School of Artificial Intelligence, Beijing University of Posts and Telecommunications \\
    \small
    $\textrm{}^{2}$State Key Laboratory of Multimedia Information Processing, School of Computer Science, Peking University\\
    \small
    $\textrm{}^{3}$National Engineering Research Center of Visual Technology, School of Computer Science, Peking University \\
    \small
    $\textrm{}^{4}$Beijing Academy of Artificial Intelligence \quad
    \small
    $\textrm{}^{5}$Tencent \\
    \small
    $\textrm{}^{6}$Beijing Key Laboratory of Multimodal Data Intelligent Perception and Governance \\
    {\tt\footnotesize $\{$zhangjingzhe,pxzhang,lisi$\}$@bupt.edu.cn} \quad {\tt\footnotesize $\{$shuchenweng,shiboxin$\}$@pku.edu.cn}
}

\newcommand\blfootnote[1]{%
  \begingroup
  \renewcommand\thefootnote{}\footnote{#1}%
  \addtocounter{footnote}{-1}%
  \endgroup
}

\begin{document}

\twocolumn[{%
\renewcommand\twocolumn[1][]{#1}%
\maketitle
\begin{center}
    \vspace{-6mm}
    \centering
    \captionsetup{type=figure}
    \includegraphics[width=\textwidth]{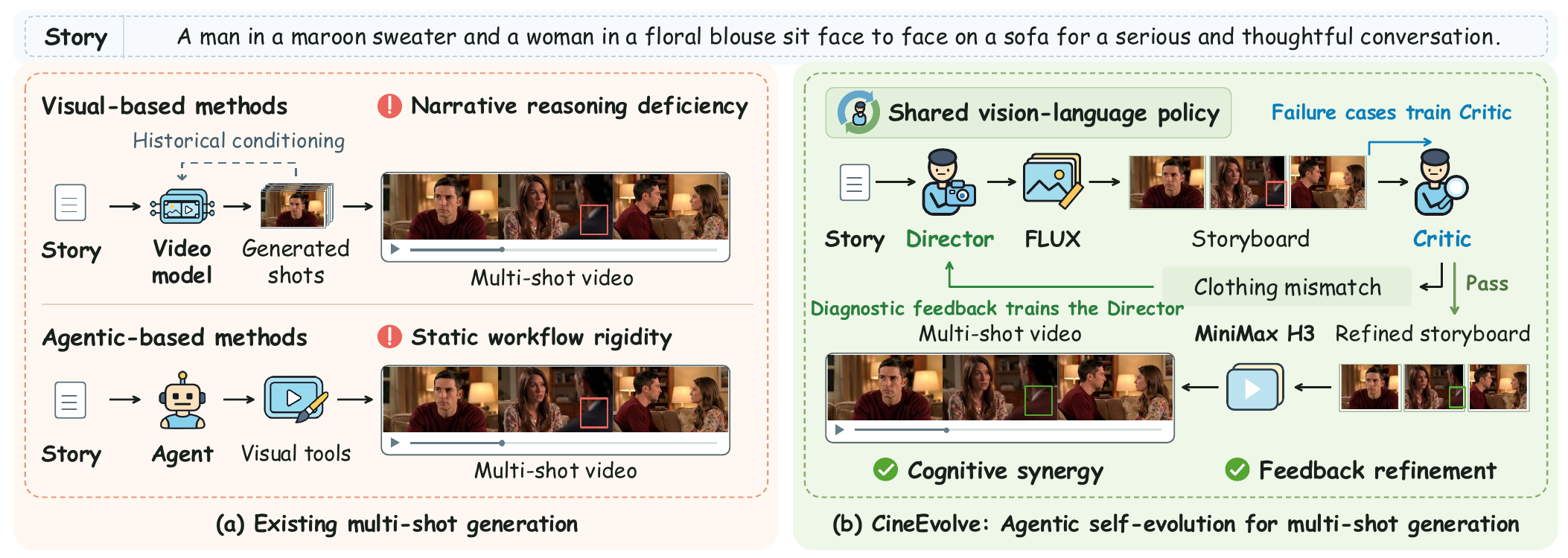}
    \captionof{figure}{
    \textbf{Illustration of our proposed CineEvolve for multi-shot video generation.}
    \textbf{(a)} Existing visual-based and agentic-based methods suffer from narrative reasoning deficiency and static workflow rigidity.
    \textbf{(b)} CineEvolve co-evolves Director and Critic in a shared vision-language policy, enabling cognitive synergy and feedback refinement.
    }
    \label{fig:teaser}
    \vspace{1mm}
\end{center}
}]
\begin{abstract} 
\blfootnote{$^\#$ Equal contributions. * Corresponding author.}
Cinematic multi-shot video generation requires both narrative planning and iterative refinement. 
Existing methods either lack narrative reasoning or rely on feed-forward workflows without policy updates. 
In this paper, we propose \textbf{CineEvolve}, an agentic self-evolution framework for cinematic multi-shot video generation that unifies a Director and a Critic within a single vision-language policy. 
Emulating real-world filmmaking that couples iterative direction with critical feedback, we train CineEvolve via alternating reinforcement learning, where the Director surfaces flaws to sharpen the Critic, and the updated Critic guides the Director's targeted revisions. 
To stabilize co-evolution, we introduce Evolutionary GRPO (E-GRPO), leveraging historical anchoring and revision safeguarding to mitigate advantage inflation. 
Additionally, we contribute \textbf{CineTraj}, a 40K cinematic trajectory dataset spanning creation, critique, and revision. 
Extensive experiments demonstrate that CineEvolve outperforms existing approaches.
Our code will be available at \href{https://github.com/JingZhe-8/CineEvolve}{this url}.

\end{abstract}

\section{Introduction}
\label{sec:introduction}

Video generation is rapidly advancing from isolated single clips toward long-horizon cinematic storytelling~\cite{rw_holocine}.
In visual storytelling, narratives unfold through sequences of interconnected shots capturing dynamic events over shifting viewpoints and scenes.
Such multi-shot sequences demand cross-shot consistency to preserve characters and environments, alongside inter-shot coherence to ensure narrative progression~\cite{rw_onestory,rw_stage}.
Satisfying these constraints requires coupling narrative planning with visual synthesis.

To address these challenges, existing visual-based methods use video generation models to produce shots, relying on extended attention mechanisms or historical conditioning~\cite{rw_cinetrans,rw_multishotmaster,rw_shotstream}.
In contrast, agentic-based methods use multimodal foundation models to plan the story structure and invoke external visual tools to render each shot~\cite{rw_vimax,rw_filmaster}.
Despite these advancements, visual-based methods lack the narrative reasoning to rectify semantic errors, while agentic-based methods typically follow feed-forward workflows or use feedback for inference-time refinement, without updating their policies through training (Fig.~\ref{fig:teaser} (a)).

To overcome this static limitation, a natural solution is to emulate real-world filmmaking, where directors iteratively refine defective shots through trial-and-error reflection.
However, visual closed-loop refinement poses practical challenges: external tools may subtly deviate from the narrative plan, and separate evaluators often struggle to assess these flaws without direct access to the creator's intent.
Because cinematic planning and defect diagnosis share an understanding of cinematic grammar, we formulate multi-shot creation as an agentic self-evolution process with two roles: a Director for initial creation and targeted revision, and a Critic for defect diagnosis.
By unifying both roles within a single vision-language policy, the model can introspect against its own narrative goals, fostering \emph{cognitive synergy} across both capabilities while enabling systematic \emph{feedback refinement} to perfect multi-shot narratives (Fig.~\ref{fig:teaser} (b)).

In this paper, we present \textbf{CineEvolve}, an agentic self-evolution framework that uses external visual generation tools for cinematic multi-shot video synthesis through a unified vision-language policy acting as both Director and Critic.
Following the supervised warm-up, the unified policy evolves through alternating reinforcement learning, where the Director exposes failure cases to train the Critic, while the updated Critic provides diagnostic feedback to guide the Director.
Specifically, the Critic is optimized for dynamic storyboard rollouts using composite critic rewards to sharpen its diagnostic precision.
Meanwhile, the Director is trained across initial creation and targeted revision, guided by the updated Critic's self-evaluative quality scores alongside narrative coherence and tool-executability rewards.
To mitigate intra-group advantage inflation during co-evolution, we propose Evolutionary GRPO (E-GRPO), which calibrates advantage estimation via revision safeguarding and historical anchoring to ensure steady improvement.

To support training and evaluation, we curate \textbf{CineTraj}, a cinematic trajectory dataset spanning initial creation, defect critique, and targeted revision, yielding 30K samples for supervised warm-up and 10K for alternating reinforcement learning.
Our contributions are summarized as follows:

\begin{itemize}
    \item We propose \textbf{CineEvolve}, an agentic self-evolution framework that unifies the Director and Critic in a single policy, driving multi-shot video generation via co-evolution.

    \item We design critic and director rewards for closed-loop optimization, alongside E-GRPO to mitigate advantage inflation via revision safeguarding and historical anchoring.
    
    \item We develop a trajectory curation pipeline and construct \textbf{CineTraj}, which contains 40K trajectory-derived cinematic samples for multi-shot generation and refinement.
    
\end{itemize}

\section{Related Work}

\subsection{Multi-Shot Video Generation}
\label{sec:related_work_multishot}
Driven by the advancements of AI-assisted filmmaking, video synthesis is expanding from isolated clips toward multi-shot storytelling.
Multi-shot video generation broadly follows two paradigms: visual-based and agentic-based approaches.
Visual-based methods enforce continuity by synthesizing consecutive shots within unified spatio-temporal latents (\eg, HoloCine~\cite{rw_holocine}, CineTrans~\cite{rw_cinetrans}, MultiShotMaster~\cite{rw_multishotmaster}) or conditioning on preceding visual frames (\eg, OneStory~\cite{rw_onestory}, ShotStream~\cite{rw_shotstream}).
While preserving visual persistence, they struggle to diagnose or rectify semantic narrative flaws.
Conversely, agentic methods leverage multimodal foundation models to schedule generative tools via structured planning (\eg, STAGE~\cite{rw_stage}, ViMax~\cite{rw_vimax}) or cinematic retrieval (\eg, FilMaster~\cite{rw_filmaster}).
However, their feed-forward execution cannot adaptively rectify execution failures.
Although CineCrew~\cite{rw_cinecrew} introduces diagnostic feedback for rule-based refinement, it lacks policy-level learning to evolve from errors.
In contrast, CineEvolve formalizes multi-shot synthesis as a closed-loop trial-and-error process, unifying Director and Critic within a single policy that co-evolves via reinforcement learning.

\subsection{Agentic Self-Evolution Mechanisms}
\label{sec:related_work_evolution}
Agentic self-evolution empowers autonomous agents to continuously improve their capabilities via experiential trial-and-error and environmental feedback.
The vast majority of existing works confine self-evolution to non-visual domains, developing reflective strategies for textual reasoning, symbolic manipulation, and algorithmic tasks through external memory updates (\eg, ReasoningBank~\cite{rw_reasoningbank}, ACE~\cite{rw_ace}, RMM~\cite{rw_rmm}), prompt optimization (\eg, GEPA~\cite{rw_gepa}, ERM~\cite{rw_erm}), self-improving code synthesis (\eg, Darwin G\"odel Machine~\cite{rw_dgm}, G\"odel Agent~\cite{rw_godel_agent}), or synthetic adversarial reasoning (\eg, R-Zero~\cite{rw_rzero}, PasoDoble~\cite{rw_pasodoble}).
Recently, a handful of pioneering studies have begun exploring self-evolution in multimodal settings; however, these efforts remain largely limited to isolated 2D perception or static image interactions (\eg, VisPlay~\cite{rw_visplay}, Agent0-VL~\cite{rw_agent0vl}, ECHO~\cite{rw_echo}, JarvisEvo~\cite{rw_jarvisevo}).
To our knowledge, CineEvolve represents the first endeavor to extend agentic self-evolution into multi-shot video generation, coupling creative direction and diagnostic critique to achieve iterative self-improvement in complex cinematic synthesis.

\section{Dataset}
\label{sec:cinevolve_dataset}

\begin{figure}[t]
    \centering
    \includegraphics[width=\linewidth]{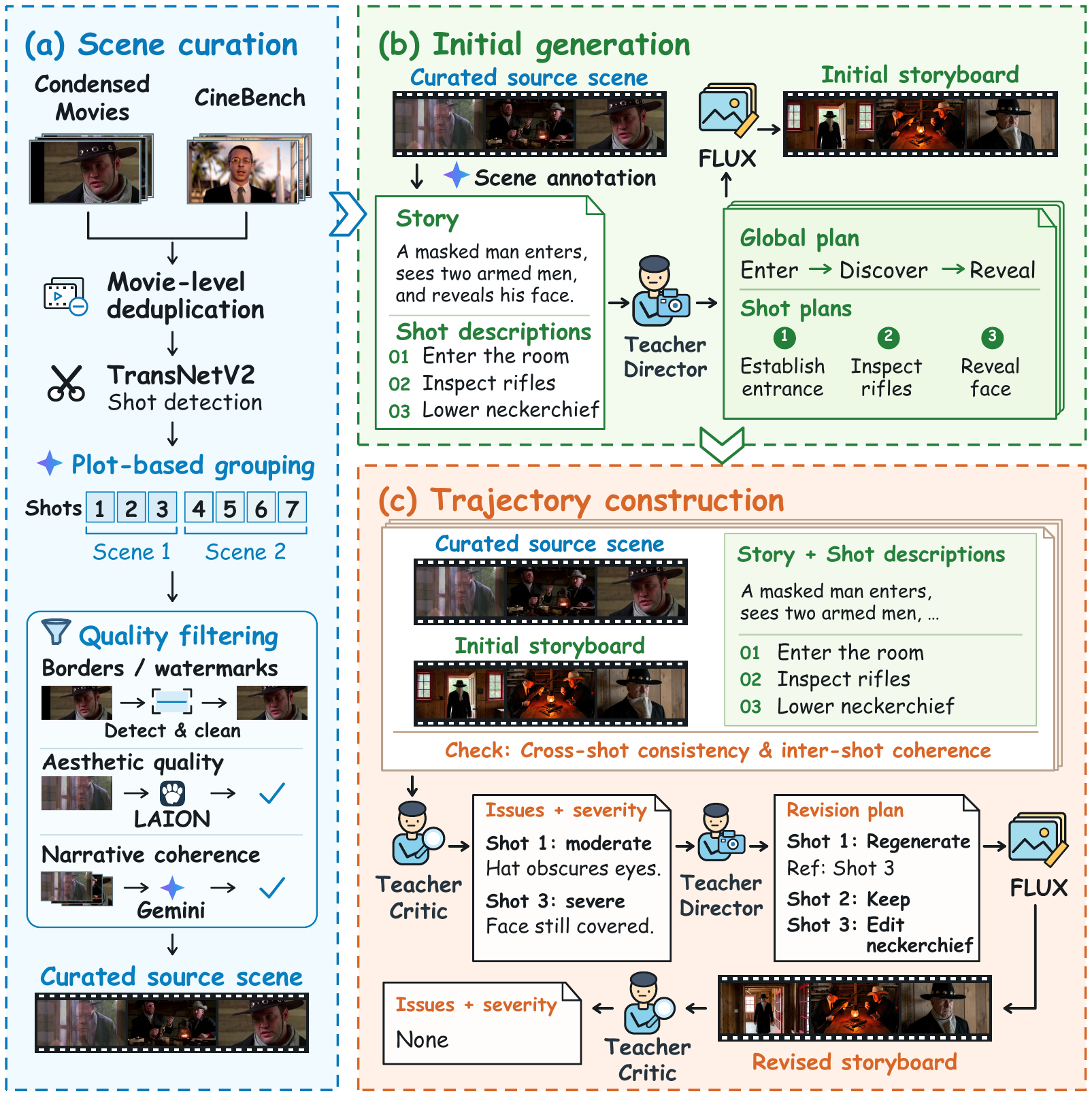}
    \caption{\textbf{Overview of the CineTraj construction pipeline.}
    \textbf{(a)} Movie scenes are curated through deduplication, shot segmentation and grouping, and quality filtering.
    \textbf{(b)} The teacher Director converts annotated stories and shot descriptions into production plans and initial storyboards.
    \textbf{(c)} Teacher-guided critique, revision, and re-evaluation produce closed-loop trajectories.}
    \label{fig:cinevolve_dataset}
    \vspace{-4mm}
\end{figure}

Generating multi-shot videos demands global visual planning and reflection across shot transitions.
Yet, existing datasets predominantly provide text-to-video pairs, lacking corrective demonstrations that are critical for models to autonomously plan and revise.
To address this limitation, we propose CineTraj, a dataset that captures the trajectory of multi-shot video generation (Fig.~\ref{fig:cinevolve_dataset}).
We use Gemini 3.1 Pro~\cite{method_gemini31pro} as the teacher VLM, Director, and Critic, with the teacher Critic also annotating refreshed replay data.

\noindent \textbf{Scene curation.} 
To construct high-quality scenes with coherent narrative arcs, we collect cinematic footage from Condensed Movies~\cite{method_cmd} and CineBench~\cite{method_cinebench} via a four-step curation pipeline:
\textit{(i) Deduplication:} performing movie-level filtering to eliminate overlapping titles between the two sources;
\textit{(ii) Boundary detection:} segmenting shot transitions across videos using TransNetV2~\cite{method_transnetv2}; 
\textit{(iii) Shot grouping:} grouping consecutive shots under native scene boundaries via VLM-assessed narrative continuity to preserve complete context; and
\textit{(iv) Quality screening:} removing black borders and watermarks~\cite{method_wan}, filtering out clips with low LAION aesthetic scores~\cite{method_laion_aesthetics}, and pruning incoherent shot sequences via Gemini-3.1-Pro~\cite{method_gemini31pro}.
This yields a high-fidelity corpus of verified source scenes with coherent visual storytelling.

\noindent \textbf{Initial generation.}
To bridge narrative intent with visual layout, we construct initial generation trajectories by transforming curated source scenes into structured production plans and visual storyboards.
Taking each curated scene $S$ as input, a teacher VLM first annotates a global narrative prompt $x$ alongside ordered shot descriptions $\mathcal{D}_{\mathrm{shot}}$ detailing visual events and character dynamics.
Conditioned on $(x,\mathcal{D}_{\mathrm{shot}})$, a teacher Director then generates a hierarchical plan $P$:
\textit{(i)} a global plan outlining narrative progression; and 
\textit{(ii)} shot-level specifications designating camera framing, subject attributes, and tool-call prompts.
Subsequently, an external visual generation model~\cite{method_flux2klein} executes these tool calls to render the multi-shot storyboard frames $B = (b_j)_{j=1}^N$, where $N$ is the number of shots.
This process outputs structured tuples $(x, P, B)$, which serve as foundational demonstrations for defect critique and targeted revision.

\noindent \textbf{Trajectory construction.}
To supervise defect diagnosis and iterative refinement, we synthesize verified closed-loop trajectories by pairing generation rollouts with teacher-guided critique and correction.
Given the source scene $S$, shot descriptions $\mathcal{D}_{\mathrm{shot}}$, and initial generation tuple $(x, P, B)$, this stage proceeds in three steps:
\textit{(i) Defect diagnosis:} the teacher Critic compares the initial storyboard $B$ against $S$ given $(x,\mathcal{D}_{\mathrm{shot}})$, focusing on cross-shot consistency and inter-shot coherence, and outputs a structured diagnosis $c = (\mathcal{P}, \mathbf{d}, \{\hat{\ell}_j\}_{j\in\mathcal{P}})$ that details shot-level defect locations, textual issues, and severity ratings (\ie, minor, moderate, or severe);
\textit{(ii) Directorial revision:} conditioned on $(x, P, B, c)$, the teacher Director formulates a targeted revision plan $P'$ with shot-level actions (\ie, keep, regenerate, or edit) and executes visual tool calls to render the revised storyboard $B'$; and
\textit{(iii) Outcome verification:} the teacher Critic re-evaluates the entire revised storyboard $B'$. We retain the demonstration only if its severity-weighted defect score strictly decreases, discarding failed or regressive attempts.
Consequently, this process outputs successful closed-loop trajectory instances $(x, P, B, c, P', B')$, which populate the demonstration corpus CineTraj.

\noindent \textbf{Data Summary.}
The constructed closed-loop trajectory dataset CineTraj provides structured supervision across three functional training objectives:
\textit{(i)} defect critique instances $(x, B, c)$ for training the Critic to predict shot defects and severity ratings; 
\textit{(ii)} initial creation instances $(x, P, B)$ for supervising directorial global planning and tool execution; and 
\textit{(iii)} targeted revision instances $(x, P, B, c, P', B')$ for training directorial defect-driven repair. 
We balance these three tasks with a sample distribution ratio of $4:2:4$, placing greater emphasis on critique and refinement.
In total, CineTraj comprises 30,000 verified trajectories for supervised policy warm-up and 10,000 diverse trajectories curated for alternating reinforcement learning\footnote{Dataset samples are provided in the supplementary materials.}.
Furthermore, we curate an independent benchmark of 100 manually verified movie sequences with strong narrative coherence, strictly excluded from all training data.

\section{Method}
\label{sec:method}

This section begins with an introduction to the training and inference paradigm (Sec.~\ref{sec:training_paradigm}). 
Then, we detail the Critic policy optimization to provide fine-grained visual critique via composite critic rewards (Sec.~\ref{sec:critic_optimization}).
Next, we elaborate on the Director policy optimization to guide generative planning via multimodal director rewards (Sec.~\ref{sec:director_optimization}).
Finally, we present Evolutionary GRPO (E-GRPO) to stabilize alternating updates and promote mutual improvement between the Director and Critic (Sec.~\ref{sec:e_grpo}).

\subsection{Training and Inference Paradigm}
\label{sec:training_paradigm}

\noindent\textbf{Stage I: Supervised Policy Warm-up.}
Directly applying reinforcement learning across multi-shot planning and diagnosis yields intractable action exploration and sparse rewards.
To establish a stable behavioral prior, we cold-start the student policy $\pi_\theta$ on CineTraj, which contains expert demonstration trajectories collected from the proprietary teacher Director and Critic.
The supervised warm-up objective is defined as:
\begin{equation}
\mathcal{L}_{\mathrm{SFT}}(\theta)
=
-\mathbb{E}_{(z,y)\sim\mathcal{D}_{\mathrm{SFT}}}
\left[\log \pi_\theta(y\mid z)\right],
\label{eq:sft_objective}
\end{equation}
where $\mathcal{D}_{\mathrm{SFT}}$ denotes the SFT demonstration subset of CineTraj, and $(z,y)$ represents a role-conditioned multimodal input and its corresponding ground-truth response.
This stage yields the warm-up policy checkpoint $\theta_0$ for subsequent evolutionary training.

\noindent\textbf{Stage II: Alternating Reinforcement Learning.}
Starting from $\theta_0$, we co-evolve the student Critic and Director across $K$ alternating rounds to transcend teacher imitation. 
In round $k\in\{0,\ldots,K-1\}$, starting from $\theta_k$, $\pi_\theta$ is first optimized as the Critic via critic rewards, yielding an intermediate checkpoint $\theta_{k,\mathrm{C}}$.
A frozen snapshot of $\theta_{k,\mathrm{C}}$ then serves as an internal evaluator, providing self-evaluation rewards that are combined with external narrative coherence rewards to update $\pi_\theta$ as the Director, producing $\theta_{k+1}$. 
This interactive co-training under a shared policy $\pi_\theta$ induces cognitive synergy: the Critic's sharpened diagnosis informs and regularizes the Director's generative planning.
By feeding the updated Director's rollouts back into $\mathcal{D}_{\mathrm{C},k+1}$ as dynamic diagnostic environments for the subsequent round, this paradigm forms an evolutionary flywheel.

\noindent\textbf{Inference.}
At test time, the autonomous policy $\pi_\theta$ executes an iterative generate-critique-revise loop without external guidance. 
Given a narrative prompt $x$, $\pi_\theta$ first acts as the Director to formulate an initial multi-shot storyboard. 
It then switches to the Critic to diagnose this storyboard. 
Upon diagnosing defects, $\pi_\theta$ alternates back as the Director to perform targeted shot-level revisions, iterating up to a maximum budget of $T_{\max}=10$ revision steps.
Once approved, the final storyboard conditions the downstream video backbone to render the cinematic multi-shot video.

\begin{figure*}[t]
    \centering
    \includegraphics[width=\linewidth]{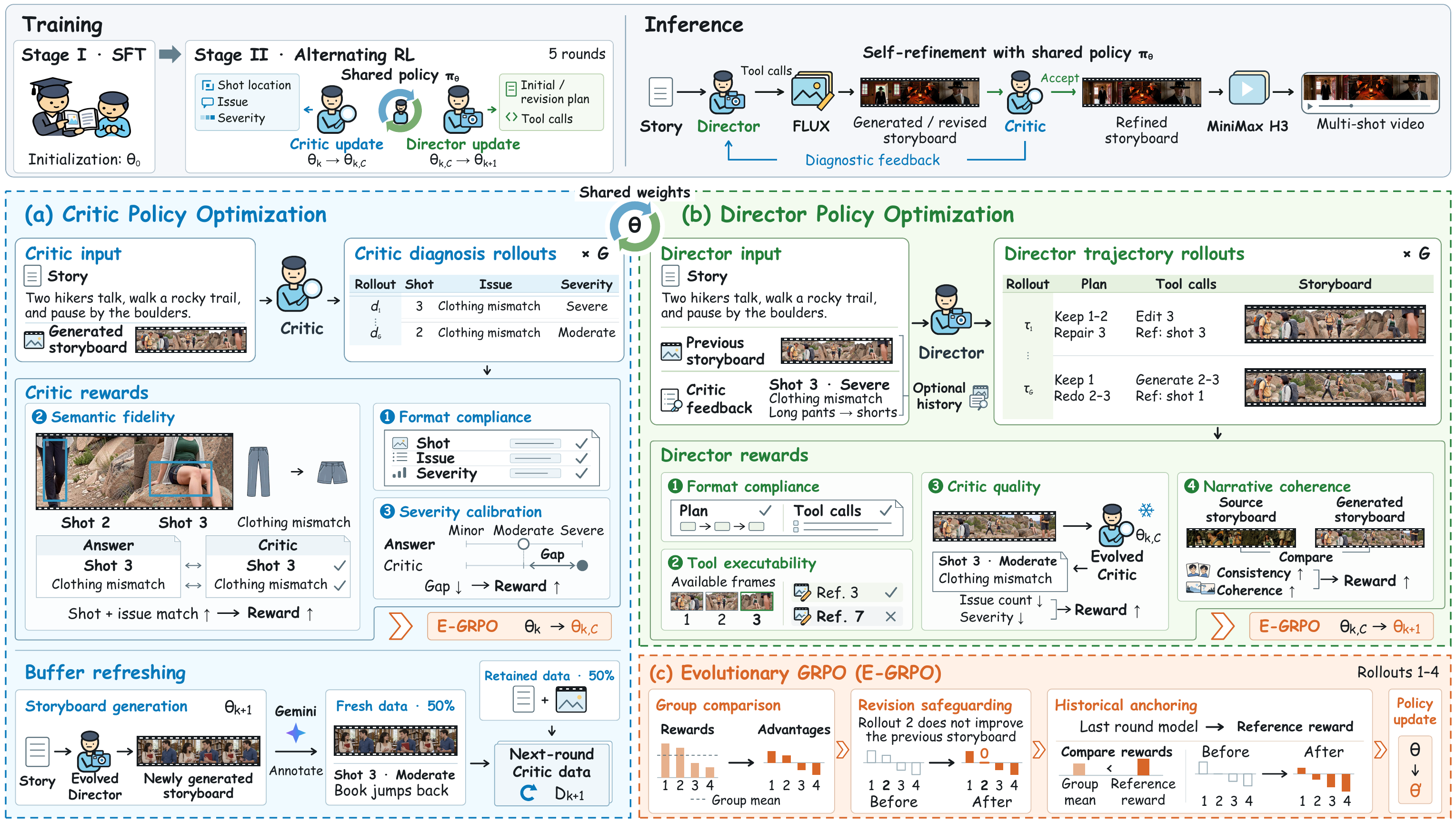}
    \caption{
        \textbf{Overview of our proposed CineEvolve framework.}
        \textbf{(a)} The Critic optimizes its diagnostic accuracy against reference defect annotations on an evolving pool of Director-generated storyboards.
        \textbf{(b)} The Director learns to draft and revise storyboards guided by self-evaluation rewards from the updated Critic, alongside narrative and visual feedback.
        \textbf{(c)} Revision safeguarding and historical anchoring are integrated to stabilize policy updates and prevent performance collapse during co-evolution.}
    \label{fig:cinevolve_method}
\end{figure*}

\subsection{Critic Policy Optimization}
\label{sec:critic_optimization}
Critic policy optimization aims to sharpen the shared policy under its Critic role, enabling it to deliver fine-grained visual critiques rather than scalar ratings (Fig.~\ref{fig:cinevolve_method} (a)).

\noindent\textbf{Critic formulation.}
Given the narrative prompt $x$ and the candidate storyboard comprising rendered visual frames $B = (b_j)_{j=1}^N$, the Critic evaluates cross-shot consistency and narrative coherence.
The Critic output is structured into three components: defective shot indices $\mathcal{P}$, corresponding defect descriptions $\mathbf{d}$, and severity ratings $\hat{\ell}_j \in \{\texttt{minor}, \texttt{moderate}, \texttt{severe}\}$ for each defective shot. 
For defect-free storyboards, the Critic outputs \texttt{none} for both defect claims and severity, yielding an empty defective set $\mathcal{P} = \emptyset$.
During training, we draw $G=4$ Critic rollouts for each sample in the replay pool $\mathcal{D}_{\mathrm{C},k}$, which is initialized from CineTraj, and compute learning signals against teacher annotations.

\noindent\textbf{Critic rewards.}
To jointly encourage structural compliance, semantic fidelity, and severity calibration, we formulate a composite critic reward across three complementary objectives\footnote{Detailed critic rewards are provided in the supplementary materials.}: 
\textit{(i) Format compliance} $F_{\mathrm{C}} \in \{0, 1\}$ enforces output schema validity, where any parsing failure invalidates the rollout and directly zeroes out the reward ($r_{\mathrm{C}} = 0$); 
\textit{(ii) Semantic fidelity} $M \in [0, 1]$ uses an F1-style formulation to penalize both false alarms and missed shots, scoring how accurately predicted defect descriptions match reference semantic units via an LLM judge on correctly located shots; and 
\textit{(iii) Severity calibration} $V \in [0, 1]$ measures agreement on defect severity levels, strictly conditioned on a high-fidelity subset of correctly identified shots whose descriptions faithfully cover reference units without spurious claims. 
Therefore, for valid rollouts, the overall critic reward is formulated as:
\begin{equation}
r_{\mathrm{C}} = 0.10 F_{\mathrm{C}} + 0.75 M + 0.15 V.
\label{eq:critic_reward}
\end{equation}

\noindent\textbf{Buffer refreshing.}
As the Director policy advances, its failure distributions shift, making static diagnostic data unrepresentative of its failure patterns.
To track these emerging weaknesses, we dynamically refresh the replay pool $\mathcal{D}_{\mathrm{C},k}$ across rounds.
After each Director optimization phase, the updated policy generates fresh storyboards, which are diagnosed by the external teacher model. 
We randomly retain 50\% of the trajectories from the previous pool $\mathcal{D}_{\mathrm{C},k}$ and add 50\% newly generated samples to form $\mathcal{D}_{\mathrm{C},k+1}$.
This refreshing mechanism maintains a fixed pool capacity while adapting the Critic to the Director's latest failure modes.

\subsection{Director Policy Optimization}
\label{sec:director_optimization}
Director policy optimization aims to train the shared policy under its Director role to generate coherent storyboards based on diagnostic feedback, using multimodal director rewards (Fig.~\ref{fig:cinevolve_method} (b)).

\noindent\textbf{Director formulation.}
We jointly optimize the shared policy for initial creation and targeted revision, as both tasks unify cinematic planning with tool execution. 
Given a narrative prompt $x$, initial creation generates a global plan with shot-level tool specifications. 
For iterative revision, the policy conditions on the story $x$, rendered storyboard $B$, and Critic diagnosis $c$ to output a revision plan and actions that selectively keep, regenerate, or edit shots.
In both tasks, a frozen visual backbone executes the dispatched generation and editing calls. Within round $k$, policy optimization follows a two-stage sequential schedule starting from $\theta_{k,\mathrm{C}}$. 
The policy is first updated on initial creation; an intermediate frozen checkpoint of this updated policy then synthesizes parent storyboards to formulate the revision task, further optimizing the same policy into $\theta_{k+1}$. 
Throughout this process, the frozen snapshot $\theta_{k,\mathrm{C}}$ serves a dual role: it provides textual diagnoses to guide revision rollouts and self-evaluation rewards for $G=4$ sampled trajectories per input.

\noindent\textbf{Director rewards.}
To balance action executability with visual narrative quality, we formulate a composite multimodal director reward. 
Specifically, the reward unifies structural validity and visual storytelling across four complementary terms\footnote{Detailed director rewards are provided in the supplementary materials.}: 
\textit{(i) Format compliance} $F_{\mathrm{D}} \in \{0, 1\}$ checks adherence to the structured output schema; 
\textit{(ii) Tool executability} $T \in \{0, 1\}$ verifies tool-call executability and image-reference validity, where violating either $F_{\mathrm{D}}$ or $T$ invalidates the rollout and yields zero reward ($r_{\mathrm{D}} = 0$); 
\textit{(iii) Critic quality} $Q \in [0, 1]$ uses the frozen Critic snapshot $\theta_{k,\mathrm{C}}$ to assess shot-level visual defect severities, where globally re-evaluating all shots during revision prevents repairing local flaws at the cost of overall quality; and 
\textit{(iv) Narrative coherence} $H \in [0, 1]$ employs an external VLM judge to score cross-shot consistency and inter-shot coherence against a curated source scene $S$, which serves purely as a narrative anchor and is strictly excluded from policy inputs.

\subsection{Evolutionary GRPO}
\label{sec:e_grpo}

Standard GRPO~\cite{method_grpo} estimates advantages through intra-group normalization, without explicitly referencing parent-storyboard or historical policy performance.
However, in alternating co-evolution, this relative comparison allows an inferior rollout to receive a falsely positive advantage whenever the entire group degrades, reinforcing counterproductive revisions.
To mitigate intra-group advantage inflation, we propose Evolutionary GRPO, which equips GRPO with revision safeguarding and historical anchoring (Fig.~\ref{fig:cinevolve_method} (c)).

\noindent\textbf{Revision safeguarding.}
To prevent the policy from adopting destructive edits, each rollout is benchmarked against its parent storyboard reward $p_g$.
At the reward level, any edit introducing new Critic-verified severe defects has its reward $r_i$ capped at $p_g$.
For each rollout group $g$, given the resulting rewards $\{r_i\}_{i=1}^G$, we compute the group advantage $a_i = (r_i - \mu_g) / \max(\sigma_g,\epsilon_{\mathrm{std}})$, where $\mu_g$ and $\sigma_g$ denote the empirical mean and standard deviation of $\{r_i\}_{i=1}^G$, and $\epsilon_{\mathrm{std}}=0.05$ is the standard-deviation floor.
However, to ensure that sub-par edits never receive positive learning signals even if they rank high in a low-quality group, we enforce an advantage safeguard:
for revision rollouts with $r_i \le p_g$, we set $\widetilde{a}_i=\min(a_i,0)$; otherwise, $\widetilde{a}_i=a_i$.

\noindent\textbf{Historical anchoring.}
To encourage steady improvement across rounds, we anchor advantages to historical performance rather than relying solely on within-group relative comparisons.
For each task (initial creation, defect critique, and targeted revision), we sample one response per input on 256 fixed calibration inputs and take the mean reward under the current evaluator as $b_k$.
Setting $p_g = 0$ for non-revision tasks, we formulate the anchored advantage as $u_i = (r_i - \max(\mu_g, b_k, p_g)) / \max(\sigma_g, \epsilon_{\mathrm{std}})$ and compute the final advantage via bounded interpolation:
\begin{equation}
A_i = (1 - \alpha) \widetilde{a}_i + \alpha u_i,
\label{eq:history_correction}
\end{equation}
where $\alpha \in [0,1]$ is the largest value retaining at least half the positive advantage mass and at most doubling the negative advantage magnitude relative to $\widetilde{a}_i$, weighted by trainable token counts over valid responses in the full optimization batch.

\begin{figure*}[p]
    \centering
    \includegraphics[width=\textwidth]{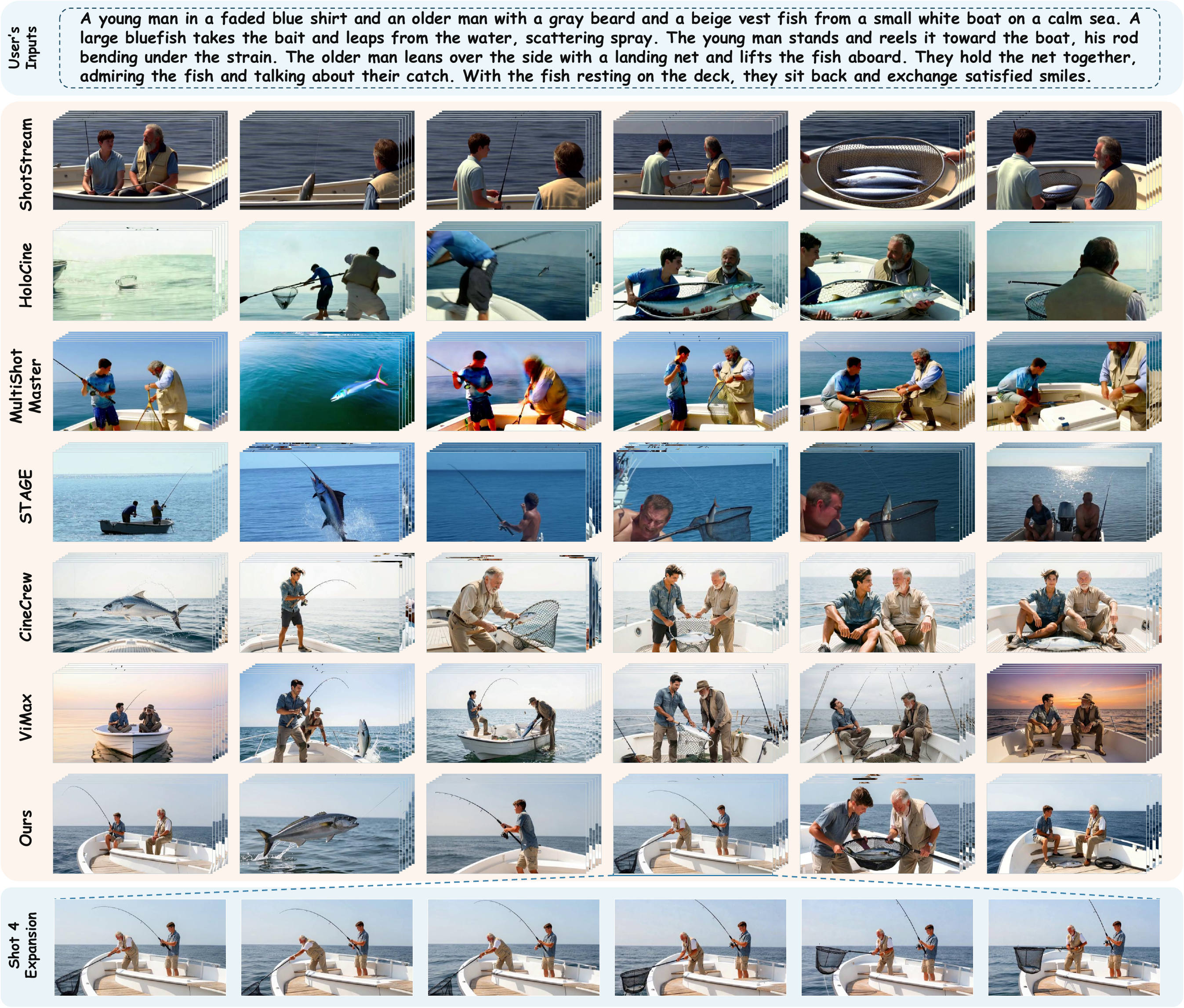}
    \caption{Qualitative comparisons with existing multi-shot video generation methods.}
    \label{fig:compare}

    \vfill 

    \captionof{table}{Quantitative comparisons and ablation results.
    $\uparrow$ ($\downarrow$) means higher (lower) is better.
    The best performances are highlighted in \textbf{bold}.}
    \label{tab:comparison_ablation}
    \small
    \setlength{\tabcolsep}{4pt}
    \begin{tabular*}{\textwidth}{@{\extracolsep{\fill}}l*{9}{c}@{}}
        \toprule
        Method
        & AQ $\uparrow$
        & IQ $\uparrow$
        & OC $\uparrow$
        & SC $\uparrow$
        & BC $\uparrow$
        & EAC $\uparrow$
        & TCA $\uparrow$
        & CSC $\uparrow$
        & TLC $\uparrow$ \\
        \midrule
        \multicolumn{10}{l}{\textit{Comparison with existing methods}} \\
        ShotStream
        & 0.5789 & 0.6881 & 0.2393 & 0.9040 & 0.9279
        & 0.6889 & 0.8144 & 0.8467 & 0.7966 \\
        STAGE
        & 0.5859 & 0.6805 & 0.2522 & 0.9285 & 0.9189
        & 0.5906 & 0.8335 & 0.7600 & 0.7309 \\
        CineCrew
        & 0.6250 & 0.7022 & 0.2469 & 0.8596 & 0.9092
        & 0.7296 & 0.7990 & 0.8013 & 0.7943 \\
        HoloCine
        & 0.5398 & 0.5612 & 0.2212 & 0.9118 & 0.9270
        & 0.5746 & 0.8156 & 0.7226 & 0.7386 \\
        ViMax
        & 0.6264 & 0.7010 & 0.2492 & 0.9322 & 0.9355
        & 0.6073 & 0.7914 & 0.8060 & 0.6939 \\
        MultiShotMaster
        & 0.5621 & 0.5764 & 0.2452 & 0.9559 & \textbf{0.9528}
        & 0.5642 & 0.8216 & 0.7833 & 0.7592 \\
        \midrule
        \multicolumn{10}{l}{\textit{Ablation studies}} \\
        SFT only
        & 0.5912 & 0.6828 & 0.2454 & 0.9299 & 0.9291
        & 0.6223 & 0.8171 & 0.7952 & 0.7529 \\
        Fixed Critic
        & 0.6093 & 0.6859 & 0.2472 & 0.9393 & 0.9320
        & 0.6782 & 0.8255 & 0.8103 & 0.7765 \\
        Separate Policies
        & 0.6185 & 0.6896 & 0.2477 & 0.9452 & 0.9400
        & 0.7040 & 0.8184 & 0.8213 & 0.7902 \\
        Original GRPO
        & 0.6206 & 0.6916 & 0.2513 & 0.9495 & 0.9421
        & 0.7179 & 0.8307 & 0.8350 & 0.7881 \\
        \midrule
        CineEvolve (Ours)
        & \textbf{0.6391}
        & \textbf{0.7155}
        & \textbf{0.2599}
        & \textbf{0.9621}
        & 0.9507
        & \textbf{0.7587}
        & \textbf{0.8456}
        & \textbf{0.8893}
        & \textbf{0.8156} \\
        \bottomrule
    \end{tabular*}
\end{figure*}

\section{Experiments}

\subsection{Training Details}
\label{sec:training_details}
We initialize the shared policy with Qwen3-VL-8B-Instruct~\cite{method_qwen3vl}, fine-tuning its language backbone while keeping the visual encoder frozen.
For generative execution, we use the four-step distilled FLUX.2-klein-9B~\cite{method_flux2klein} for storyboard generation and editing, and the eight-step distilled MiniMax-H3 Turbo~\cite{method_minimax_h3_turbo} for video generation, with both visual backbones kept frozen.
Supervised policy warm-up is conducted for two epochs with a learning rate of $10^{-5}$ and an effective batch size of 64.
Subsequently, we perform five rounds of alternating reinforcement learning, with 64 optimization steps allocated per role in each round.
We use a learning rate of $10^{-6}$ and a batch size of $16$.
All experiments are conducted on 16 NVIDIA A100 GPUs.

\begin{figure*}[!t]
    \centering
    \includegraphics[width=\textwidth]{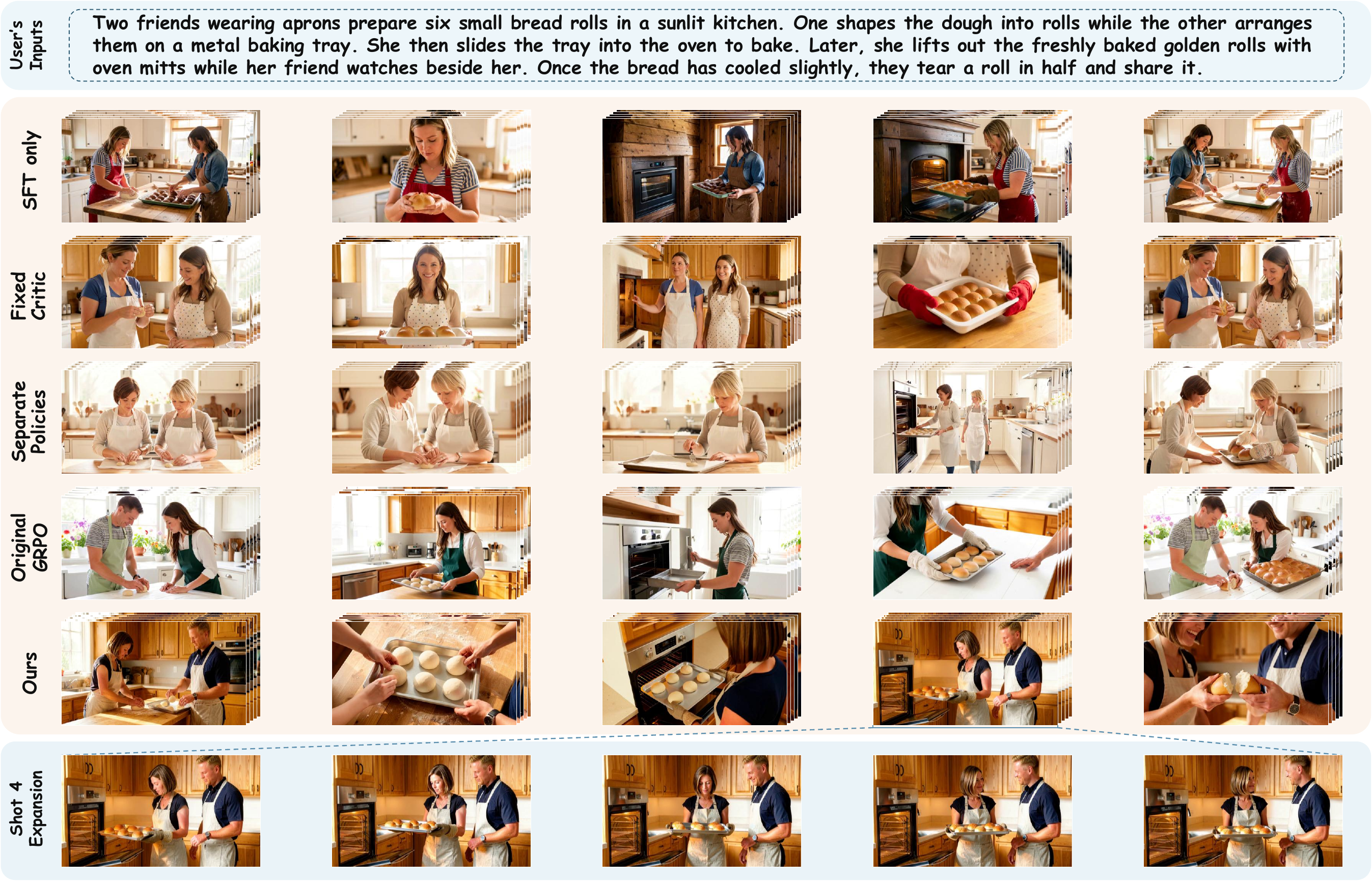}
    \caption{Ablation study results with different variants of our CineEvolve framework.}
    \label{fig:supp_ablation}
\end{figure*}

\subsection{Evaluation Metrics}
\label{sec:evaluation_metrics}

We comprehensively evaluate CineEvolve across three aspects\footnote{Metric details are provided in the supplementary materials.}: 
\textit{(i) Visual Quality and Alignment:} we assess aesthetic appeal, technical fidelity, and text and video alignment using \textbf{Aesthetic Quality (AQ)}, \textbf{Imaging Quality (IQ)}, and \textbf{Overall Consistency (OC)} metrics from VBench~\cite{eval_vbench}.
\textit{(ii) Within-Shot Consistency:}
we evaluate the temporal stability of subjects and scene backgrounds across successive frames using \textbf{Subject Consistency (SC)} and \textbf{Background Consistency (BC)} metrics from VBench~\cite{eval_vbench}.
\textit{(iii) Cross-Shot Coherence:}
we evaluate visual continuity and narrative transitions across shots through four complementary metrics. Specifically, \textbf{Entity Appearance Consistency (EAC)} and \textbf{Transition Contrast Agreement (TCA)} measure cross-shot entity similarity and visual transition fidelity using DINOv2~\cite{eval_dinov2} and CLIP~\cite{eval_clip} features. Additionally, \textbf{Cross-shot Consistency (CSC)} and \textbf{Transition Logic Consistency (TLC)} leverage a Vision-Language Model~\cite{method_gemini31pro} evaluator to assess fine-grained entity attributes and inter-shot narrative progression, referencing the original story and source scene.

\subsection{Comparison with Existing Methods}
\label{sec:comparison}

We compare CineEvolve with six representative multi-shot video generation baselines, categorized into direct visual synthesis methods (HoloCine~\cite{rw_holocine}, MultiShotMaster~\cite{rw_multishotmaster}, and ShotStream~\cite{rw_shotstream}) and agentic methods (STAGE~\cite{rw_stage}, ViMax~\cite{rw_vimax}, and CineCrew~\cite{rw_cinecrew}). All methods are evaluated on the proposed CineTraj benchmark (Sec.~\ref{sec:cinevolve_dataset}).

\noindent\textbf{Fair Comparisons.} 
To ensure a fair evaluation, all baselines share identical backbones with CineEvolve: Qwen3-VL-8B-Instruct serves as the VLM for agentic baselines, while FLUX.2-klein-9B and MiniMax-H3 Turbo are adopted as the unified image and video backbones.
For visual-based baselines, we supply detailed shot-level scripts pre-expanded by Qwen3-VL-8B-Instruct to ensure fairness.

\noindent\textbf{Qualitative Comparisons.}
In Fig.~\ref{fig:compare}, we present visual quality comparison results with the aforementioned methods.
ShotStream~\cite{rw_shotstream} fails to strictly adhere to numerical constraints specified in the prompt (\eg, \cref{fig:compare} first row, it violates the entity count by depicting multiple fish in shot~5, undermining prompt fidelity).
HoloCine~\cite{rw_holocine} suffers from poor narrative progression and pacing (\eg, \cref{fig:compare} second row, it abruptly omits the initial scene setup in shot~1, leading to a confusing storyline).
MultiShotMaster~\cite{rw_multishotmaster} exhibits notable visual artifacts and attribute instability (\eg, \cref{fig:compare} third row, the character's clothing color drastically alters alongside severe facial distortions across shot 3 and shot 4).
STAGE~\cite{rw_stage} suffers from severe attire drift across long-range transitions (\eg, \cref{fig:compare} fourth row, the protagonist's outfit inconsistently morphs between shot~1 and shot~6, breaking identity persistence).
CineCrew~\cite{rw_cinecrew} opens too abruptly without establishing adequate character context (\eg, \cref{fig:compare} fifth row, the absence of an establishing shot disrupts the cinematic pacing).
ViMax~\cite{rw_vimax} struggles with environmental consistency across continuous scenes (\eg, \cref{fig:compare} sixth row, an unnatural lighting discontinuity occurs between daylight and sunset across shot 4 to shot 6, creating an incoherent visual tone).
In contrast, CineEvolve preserves character identity and lighting consistency across all frames while depicting a coherent progression from the initial fishing scene to landing the catch and the final conversation, delivering a highly cinematic and coherent sequence.

\noindent \textbf{Quantitative Comparisons.}
We present quantitative comparisons in \cref{tab:comparison_ablation}, where CineEvolve consistently outperforms existing multi-shot generation approaches across eight metrics. 
This demonstrates its superior ability to create multi-shot cinematic videos with high aesthetic appeal and technical fidelity (AQ, IQ), faithful story-to-video alignment (OC), robust intra-shot temporal stability (SC, BC), reliable cross-shot visual persistence (EAC, CSC), and coherent narrative progression with logical transitions (TCA, TLC).

\begin{table}[t]
    \centering
    \caption{Human preference shares (\%) across four evaluation dimensions. Higher is better.}
    \label{tab:user_study}
    \small
    \setlength{\tabcolsep}{4pt}
    \begin{tabular*}{\columnwidth}{@{\extracolsep{\fill}}lcccc@{}}
        \toprule
        Method & VQ & TA & CC & IC \\
        \midrule
        ShotStream      & 7.2  & 6.6  & 6.0  & 6.6  \\
        STAGE           & 5.8  & 6.0  & 6.2  & 6.8  \\
        CineCrew        & 7.2  & 5.8  & 8.4  & 8.8  \\
        HoloCine        & 10.2 & 10.2 & 5.4  & 5.6  \\
        ViMax           & 6.8  & 7.0  & 9.6  & 9.6  \\
        MultiShotMaster & 10.6 & 10.4 & 5.8  & 5.8  \\
        \midrule
        CineEvolve (Ours)
        & \textbf{52.2}
        & \textbf{54.0}
        & \textbf{58.6}
        & \textbf{56.8} \\
        \bottomrule
    \end{tabular*}
\end{table}

\noindent\textbf{User Study.}
We evaluate human preferences across four dimensions\footnote{Details of the user study are provided in the supplementary material.}: Visual Quality (VQ), Text Alignment (TA), Cross-shot Consistency (CC), and Inter-shot Coherence (IC).
We randomly select 25 test stories and recruit 20 volunteers from Amazon Mechanical Turk (AMT) to independently select their preferred video among the seven methods for each dimension.
As shown in Table~\ref{tab:user_study}, CineEvolve receives the highest preference share in all four dimensions.

\subsection{Ablation Studies}
\label{sec:ablation}
To validate the effectiveness of our key design and training paradigms, we construct four ablation variants. The evaluation scores and visual results of the ablation study are presented in Table~\ref{tab:comparison_ablation} and Fig.~\ref{fig:supp_ablation}, respectively.

\noindent\textbf{SFT Only.}
We remove alternating reinforcement learning to retain only the supervised warm-up policy. Without iterative trial-and-error reflection, the model fails to continuously refine its directorial planning and diagnostic precision, yielding degraded performance across all metrics. 
As shown in \cref{fig:supp_ablation} (row 1), the generated sequence suffers from chronological reversals and environmental drift, where a baked roll prematurely appears in Shot~2 before the baking stage.

\noindent\textbf{Fixed Critic.}
We freeze the Critic after supervised warm-up while optimizing only the Director during reinforcement learning. Lacking dynamic diagnostic adaptation, the static Critic fails to provide informative feedback across directorial rollouts. Thus, this variant degrades cross-shot consistency (lower EAC and CSC scores). 
As shown in \cref{fig:supp_ablation} (row 2), it fails to preserve object-level consistency, resulting in changes in the count and spatial arrangement of bread rolls on the tray between Shot~2 and Shot~4.

\noindent\textbf{Separate Policies.}
We optimize the Director and Critic separately rather than within a unified model. Without unified parameter sharing, the model loses cognitive synergy in cinematic grammar, resulting in degraded transition coherence (lower TLC score). 
As shown in \cref{fig:supp_ablation} (row 3), it ambiguously conflates loading and unloading the oven in Shot~4, disrupting the transition from raw dough to baked bread.

\noindent\textbf{Original GRPO.}
We replace E-GRPO with standard GRPO, removing revision safeguarding and historical anchoring. Without baseline calibration, standard GRPO falsely rewards defective candidates that rank highest in a poor group, encouraging flawed revisions that harm entity consistency and transition logic (lower CSC and TLC scores). 
As shown in \cref{fig:supp_ablation} (row 4), it exhibits identity and state flickering, where the woman's attire abruptly changes from long to short sleeves between Shot 2 and Shot 3.

\section{Conclusion}
\label{sec:conclusion}

In this paper, we present \textbf{CineEvolve}, an agentic self-evolution framework for cinematic multi-shot video generation. 
By unifying the Director and Critic roles within a single vision-language policy, CineEvolve realizes iterative multi-shot synthesis via alternating reinforcement learning. 
To sustain stable co-evolution, we introduce role-specific rewards that couple fine-grained diagnostic fidelity with self-evaluative feedback, complemented by E-GRPO to calibrate policy optimization. 
Furthermore, we construct \textbf{CineTraj}, a large-scale cinematic trajectory dataset comprising 40K samples for cinematic multi-shot video generation. 
Extensive experiments demonstrate that CineEvolve significantly outperforms existing approaches in both narrative coherence and visual execution, establishing a promising paradigm for autonomous cinematic creation.

\section*{Limitations}
CineEvolve primarily performs defect diagnosis and targeted revision at the storyboard level.
Although multi-shot narrative consistency and cinematic composition are predominantly governed by keyframe storyboards, subtle motion artifacts can still arise during subsequent video synthesis, which remain outside the current feedback loop.
We expect these residual artifacts to diminish naturally as foundational video generation backbones continue to advance.

{\small
\bibliographystyle{ieeenat_fullname}
\bibliography{main}
}

\end{document}